\documentclass[runningheads]{llncs}

\usepackage[T1]{fontenc}
\usepackage{amsmath, amssymb, mathtools}
\usepackage{graphicx}
\usepackage{tikz}
\usepackage{booktabs, multirow}
\usepackage{arydshln}
\usepackage{wrapfig}
\usepackage{floatrow}
\usepackage[table,dvipsnames]{xcolor}
\definecolor{brightgreen}{rgb}{0.4,1.0,0.0}
\usepackage{pifont}
\usepackage{comment}
\usepackage{hyperref}
\usepackage{dsfont}   
\usepackage[capitalize,noabbrev]{cleveref}

\usepackage[textsize=tiny]{todonotes}
\begin{document}
\title{PrediTree: A Multi-Temporal Sub-meter Dataset of Multi-Spectral Imagery Aligned With Canopy Height Maps}
%
%

\author{Hiyam Debary \and Mustansar Fiaz \and Levente Klein}
\authorrunning{H. Debary et al.} 

\institute{IBM Research, \\ \email{hiyam.debary@ibm.com}, \email{mustansar.fiaz@ibm.com}, \email{kleinl@us.ibm.com}}

%
%

\maketitle              
\begin{abstract}
We present PrediTree, the first comprehensive open-source dataset designed for training and evaluating tree height prediction models at sub-meter resolution. This dataset combines very high-resolution (0.5m) LiDAR-derived canopy height maps, spatially aligned with multi-temporal and multi-spectral imagery, across diverse forest ecosystems in France, totaling 3,141,568 images. PrediTree addresses a critical gap in forest monitoring capabilities by enabling the training of deep learning methods that can predict tree growth based on multiple past observations.
To make use of this PrediTree dataset, we propose an encoder-decoder framework that requires the multi-temporal multi-spectral imagery and the relative time differences in years between the canopy height map timestamp (target) and each image acquisition date for which this framework predicts the canopy height. The conducted experiments demonstrate that a U-Net architecture trained on the PrediTree dataset provides the highest masked mean squared error of $11.78\%$, outperforming the next-best architecture, ResNet-50, by around $12\%$, and cutting the error of the same experiments but on fewer bands (red, green, blue only), by around $30\%$.
This dataset is publicly available on \href{https://huggingface.co/datasets/hiyam-d/PrediTree}{HuggingFace}, and both processing and training codebases are available on \href{URL}{GitHub}.
\end{abstract}
\section{Introduction}
Predicting tree height growth over time has long been a central focus in ecology and forestry research, offering valuable insights into carbon sequestration dynamics, forest management strategies, and climate change adaptation \cite{chave2014improved,longo2016aboveground,rohner2018predicting}. \cite{landsberg1997generalised} introduced 3-PG (Physiological Principles Predicting Growth), a physics-based model designed to simulate stand-level forest growth by leveraging monthly climate data, either observed or simulated, soil properties, and species-specific physiological parameters. 
Remotely sensed data, particularly very high-resolution imagery and canopy height maps (CHMs), offers a promising avenue for tree growth modeling \cite{tolan2024very,li2020high,lang2023high}. 
Unlike most existing CHM datasets \cite{tolan2024very,lang2023high,liu2023overlooked,allred2025canopy,schwartz2022high}, that focus on single-timestamp height estimation, PrediTree specifically addresses the task of height prediction -- forecasting future tree heights from historical observations.\\
The ambition of this paper is to establish a dataset enabling the training of tree height prediction models. Here, we refer to assessing tree height from features acquired at the same timestamp as \textbf{estimation}, and assessing tree height from features acquired at past timestamps as \textbf{prediction}. More specifically, in the context of this study, the target $Y$ is the CHM at some time $t_y$, and the input features $I_1, \dots, I_N$ are multi-spectral data, spatially aligned with the CHM, acquired at times $t_1, \dots, t_N$ with $t_1 < \dots < t_N < t_y$. The typical duration $\Delta t$ between two snapshots is between 1 and 3 years. 
To handle such a challenging dataset, we propose an encoder-decoder architecture, which inputs multi-temporal multi-spectral data and  the relative time differences in the years between the CHM timestamp $t_y$ (target) and each image acquisition date $I_1, \dots, I_N$. An experimental study demonstrates that the proposed framework can predict tree growth.
We believe that this is the first openly available dataset combining sub-meter resolution (0.5m) LiDAR-derived CHMs with multi-temporal, multi-spectral imagery across diverse forest ecosystems. 
Unlike existing datasets that focus on single-timestamp height estimation, PrediTree enables training models to predict future tree heights from historical observations. 
\begin{table}[t]
\centering
\small
\setlength{\tabcolsep}{2pt} 
\scalebox{0.6}{
\begin{tabular}{clllllrlrccclr}
\toprule
& \textbf{Author} 
& \multicolumn{4}{c}{\textbf{MS images}} 
& \multicolumn{3}{c}{\textbf{CHMs}} 
& \multicolumn{3}{c}{\textbf{OS}} 
& \multicolumn{2}{c}{\textbf{Extent}} \\
\cmidrule(lr){3-6} 
\cmidrule(lr){7-9} 
\cmidrule(lr){10-12} 
\cmidrule(lr){13-14} 
&  
& R [m] & Sensor & Timestamps & Bands 
& R [m] & Sensor & Samples 
& MS & CHM & Code 
& Zones & Area [km$^2$] \\
\midrule

\multicolumn{14}{c}{\textbf{SINGLE-TEMPORAL DATASETS}} \\
\midrule
& Tolan et al.~\cite{tolan2024very}    & 0.59 & Maxar    & 1        & RGB           & 1    & ALS+GEDI & 5.8k   & \textcolor{red}{\ding{55}} & \textcolor{brightgreen}{\checkmark} & \textcolor{red}{\ding{55}} & US, Brazil & 5.8k \\
& Lang et al.~\cite{lang2023high}      & 10   & S2       & 1        & 12              & 25   & GEDI     & 600M   & \textcolor{brightgreen}{\checkmark} & \textcolor{brightgreen}{\checkmark} & \textcolor{red}{\ding{55}} & Global     & 14k \\
& Liu et al.~\cite{liu2023overlooked} & 3    & Planet   & 1        & RGB+NIR       & 3    & ALS      & 100k   & \textcolor{red}{\ding{55}} & \textcolor{orange}{\ding{108}} & \textcolor{orange}{\ding{108}} & Europe     & 700k \\
& Allred et al.~\cite{allred2025canopy}  & 1    & NAIP     & 1        & RGB+NIR       & 1    & ALS      & 22M    & \textcolor{brightgreen}{\checkmark} & \textcolor{brightgreen}{\checkmark} & \textcolor{brightgreen}{\checkmark} & US         & ~1.5M \\
& Schwartz et al.~\cite{schwartz2022high} & 10   & S1/S2    & 1        & 10              & 25   & GEDI     & 90k    & \textcolor{red}{\ding{55}} & \textcolor{orange}{\ding{108}} & \textcolor{orange}{\ding{108}} & France     & ~13k \\

\midrule
\multicolumn{14}{c}{\textbf{MULTI-TEMPORAL DATASETS}} \\
\midrule
& Potapov et al.~\cite{potapov2021mapping}     & 30   & Landsat   & 3 (monthly) & RGB+NIR+SWIR & 25   & GEDI     & 372    & \textcolor{red}{\ding{55}} & \textcolor{orange}{\ding{108}} & \textcolor{orange}{\ding{108}} & Global     & ~150k \\
& Turubanova et al. ~\cite{turubanova2023tree}     & 30   & Landsat   & 21 (yearly) & 7              & 1–10 & ALS      & $>$1M  & \textcolor{red}{\ding{55}} & \textcolor{orange}{\ding{108}} & \textcolor{orange}{\ding{108}} & Europe     & ~5.5M \\
& Foge et al.~\cite{fogel2024open}   & 1.5  & SPOT 6/7  & 3 (yearly)  & RGB+NIR      & 1.5  & ALS      & 87k    & \textcolor{brightgreen}{\checkmark} & \textcolor{brightgreen}{\checkmark} & \textcolor{brightgreen}{\checkmark} & France     & 87k \\
\rowcolor[HTML]{F8E2C7}
& \textbf{Ours}                  & 0.5  & IGN    & 3 (yearly)  & RGB+NIR+NDVI & 0.5  & ALS      & 785k   & \textcolor{brightgreen}{\checkmark} & \textcolor{brightgreen}{\checkmark} & \textcolor{brightgreen}{\checkmark} & France     & 13k \\

\bottomrule
\end{tabular}}
\caption{Summarization of existing single-temporal and multi-temporal multi-spectral (MS) and canopy height map (CHM) datasets. For each dataset, the resolution (R) is indicated for both the multi-spectral and CHM imagery, as well as the sensors used. The open-sourceness (OS) of each of these datasets is noted. \textcolor{orange}{\ding{108}}: partially open-source.}
\label{tab:intro-table}
\vspace{-0.5em}
\end{table} 
\vspace{-1em}

\section{Related work}
Remote sensing datasets that combine multi-spectral imagery with CHMs are increasingly common and have supported a wide range of applications, from above-ground biomass estimation to forest monitoring \cite{simard2011mapping,duncanson2022aboveground,tolan2024very,webb2021vegetation,kwon2024canopy,lefsky2005estimates,lang2023high}. However, the majority of these datasets consist of imagery and CHM layers acquired at the same timestamp, making them suitable for static canopy height estimation, but unfit for tasks involving tree growth modeling across time. Relevant datasets are represented in \cref{tab:intro-table}.

\subsection{Datasets for CHM estimation}
Several datasets have been developed for estimating CHMs from remote sensing inputs collected at a single timestamp. For instance, \cite{lang2023high} generated a 10m resolution global CHM for 2020 using Sentinel-2 imagery in conjunction with GEDI waveform-derived heights. Similarly, \cite{schwartz2022high} used Sentinel-1 and Sentinel-2 data, in combination with GEDI to map canopy height in France. At higher resolution, \cite{tolan2024very} fused Maxar satellite imagery with airborne and GEDI LiDAR to produce sub-meter CHMs, while \cite{wagner2024sub} focused on CHMs across California using NAIP and 3DEP LiDAR. \cite{allred2025canopy} introduced a considerable collection of sub-meter CHMs based on NAIP imagery and USGS LiDAR across the United States, and \cite{liu2023overlooked} used 3m PlanetScope data to generate CHMs for European forests.



\begin{wrapfigure}{r}{0.5\textwidth}
\vspace{-1em} 
\centering
\includegraphics[width=\linewidth]{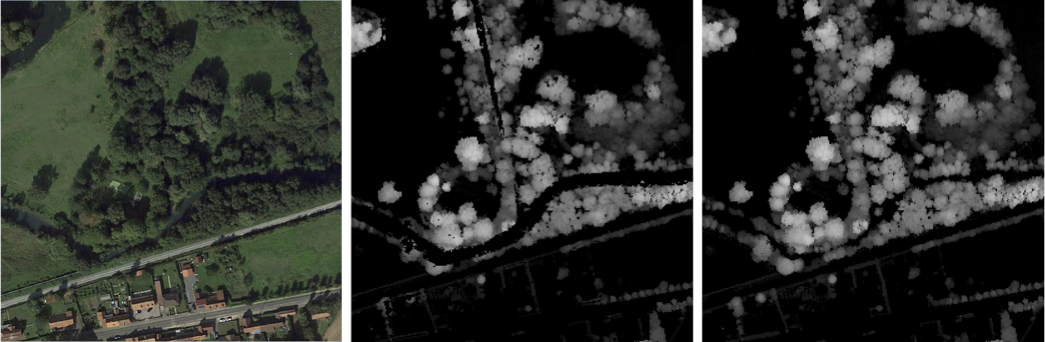}
\caption{Effect of DTM smoothing on CHM quality near water features. Left: RGB aerial imagery showing a forest area intersected by a water body (center of the image). Center: Unprocessed DTM without smoothing, exhibiting data gaps (black areas), particularly above the water body. Right: Improved DTM produced using a square filter of side 10m, effectively filling gaps where canopy information would otherwise be truncated.}
\vspace{-0.5em}
\label{fig:smoothing}
\end{wrapfigure}
\vspace{-1.0em}

\subsection{Datasets for tree growth prediction}
\cite{potapov2021mapping} used Landsat imagery across monthly timesteps and GEDI to map forest structure globally, but only at coarse 30m resolution. \cite{turubanova2023tree} tracked annual tree height changes across Europe using 21 years of Landsat data and Airborne Laser Scanning (ALS) derived CHMs, offering valuable long-term insights but still at a coarse resolution. At finer scale, \cite{fogel2024open} introduced a French dataset using SPOT 6/7 (1.5m resolution) with co-temporal Red, Green, Blue (RGB) and LiDAR imagery for three consecutive years (2021-2023), specifically for change detection purposes, making it one of the first to support high-resolution growth modeling.\\
Despite these advances, we observe that these datasets suffer from three key limitations: 1) \textbf{temporal alignment}: most datasets provide imagery and CHM from the same timestamp, preventing growth modeling, 2) \textbf{spatial resolution}: coarse resolution (10-30m) datasets cannot capture individual tree crowns, 3) \textbf{data availability}: proprietary or restricted access limits reproducibility and training models.
To address this, we present \textbf{PrediTree}, a novel sub-meter resolution dataset that includes RGB, Near Infrared (NIR), and Normalized Difference Vegetation Index (NDVI)  across three time points, that are not restricted to consecutive years and that vary for each location, and paired with CHMs from a later year. With over 3,141,568 $256\times256$ tiles at 0.5m, PrediTree is designed to support the training of deep learning models capable of learning spatio-temporal vegetation growth patterns. 


\section{Method}
\label{sec:formatting}
In this section, we present the methodology for the construction of the PrediTree benchmark dataset. 
The French national forest inventory data from IGN (Institut national de l’information géographique et forestière) provides raw classified cloud point imagery of the majority of the French territory \cite{ign}, approximately 80 departments, delivered at a single temporal snapshot and with a point density $> 10$ points/m$^2$. 
Additionally, they offer corresponding aligned optical imagery that includes RGB and NIR-RG orthophotos at various very high resolutions: 0.15m, 0.2m, or 0.5m. Since the RGB and NIR-RG tiles cover larger areas than the cloud point tiles, we identify the smaller cloud point tiles that fit within the boundaries of these optical imagery tiles.

\begin{wrapfigure}{r}{0.55\textwidth}
    \centering
    \vspace{-1.3cm}
    \includegraphics[width=\linewidth]{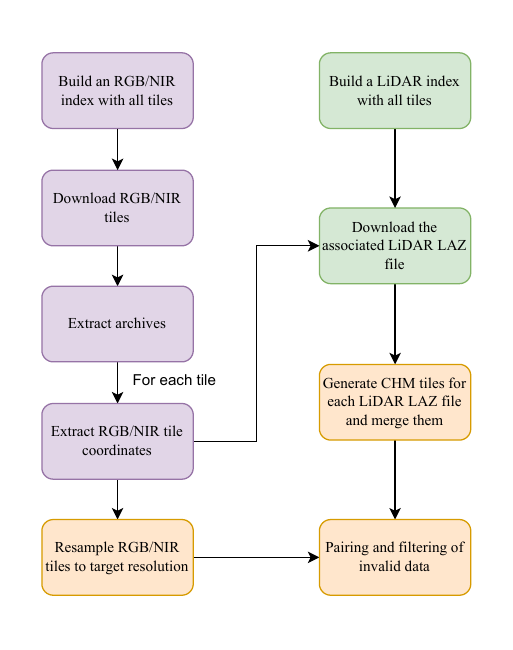}
    \caption{Flowchart of the three-phase methodology for dataset creation. \textbf{Phase 1} (purple boxes) shows the RGB and NIR-RG data acquisition process, from building an index to extracting coordinates. \textbf{Phase 2} (green boxes) depicts the LiDAR data acquisition workflow, including downloading LAZ files and generating CHM tiles. \textbf{Phase 3} (orange boxes) represents the processing, pairing, and filtering steps that integrate data from both sources into the final harmonized dataset.}
    \label{fig:flowchart}
    \vspace{-0.5cm}
\end{wrapfigure}
\begin{figure}[t]
    \centering    
    \includegraphics[width=\textwidth]{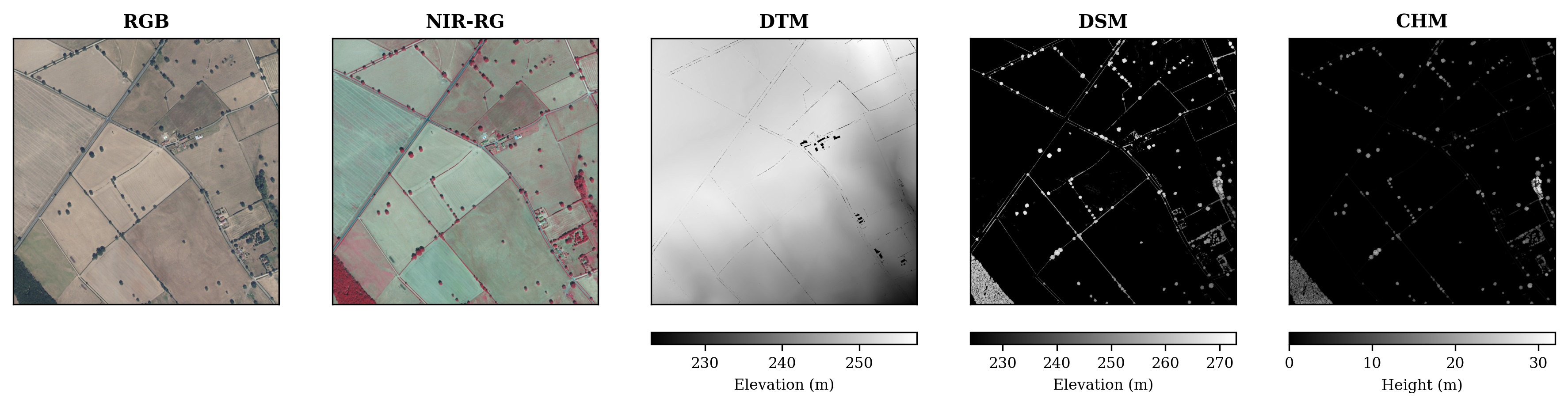}   
    \caption{Visualization of the obtained products through the described pipeline. From left to right: RGB orthophotography, fake color representation (using NIR, R and G bands), DTM, DSM, and CHM. All products displayed at the same spatial resolution of 0.5m. The elevation products use a grayscale representation with lighter tones indicating higher values.}
    \label{fig:products}
\end{figure}

This data processing and harmonization methodology follows a structured workflow consisting of several integrated phases that can be applied to any French department and scaled to handle varying spatial extents. Initially, we process both optical and LiDAR data from the IGN database. The optical data including RGB and NIR-RG are in GeoTIFF formats, while LiDAR data is obtained as classified point clouds in COPC (Cloud Optimized Point Cloud) LAZ format.\\
The first step of processing of the input data is the concatenation of the RGB and NIR bands. The last channel (NDVI) is deduced using the R and NIR bands. As for the LiDAR processing phase, we leverage the PDAL (Point Data Abstraction Library) package to process the point cloud data into a raster. 
This workflow implements processing pipelines that first separate ground points from vegetation points to generate Digital Terrain Models (DTM) and Digital Surface Models (DSM), respectively. 
A critical parameter in this methodology involves the smoothing of the DTM values by using a square filter of side 10m, which was chosen for computational efficiency purposes. This helps ensure a continuous terrain model by filling small gaps in the ground point coverage, as depicted in \cref{fig:smoothing}. 
The CHM is subsequently derived by subtracting the smoothed DTM from the DSM, effectively isolating vegetation height with respect to the ground, while minimizing data gaps.\\
The processing continues with merging multiple CHM tiles to match the spatial extent of the RGB-NIR-NDVI products. This implementation merges overlapping tiles by taking the most recent data at the overlap location. During this merging process, we extract and preserve metadata from the source tiles, including the acquisition timestamps. This allows us to calculate the mean acquisition year across all merged CHM tiles, providing an important temporal reference for the final CHM product that accurately represents the average forest conditions captured in the dataset. The optical imagery is then resampled from its native very high resolution to the target resolution, which for this study we chose to be 0.5m, using an interpolation method with the target grid matching that of the CHM data. \\ 
Finally, quality control measures are applied to filter out invalid or problematic data, with explicit masking of NaN values in the CHM and corresponding zero values in the imagery. In \cref{fig:flowchart}, we summarize the methodology to produce the harmonized optical data with their aligned CHMs, where the target resolution is configurable as an input parameter. 
For departments with more than three timestamps, we randomly choose three timestamps. After this process, we end up with 33 departments that have exactly three timestamps. \\ 
We represent in \cref{fig:products} two samples obtained through the end-to-end data processing pipeline described in this section, and provide the code used to generate it in \href{URL}{GitHub}. 
The type of all five bands is unsigned integer 8, even for NDVI -- we translate the values from $-1\rightarrow 1$ to $0 \rightarrow 255$ to maintain data-type homogeneity between all bands. 
\vspace{-0.5em}
\begin{figure}[ht!]
    \centering
    \includegraphics[width=\linewidth]{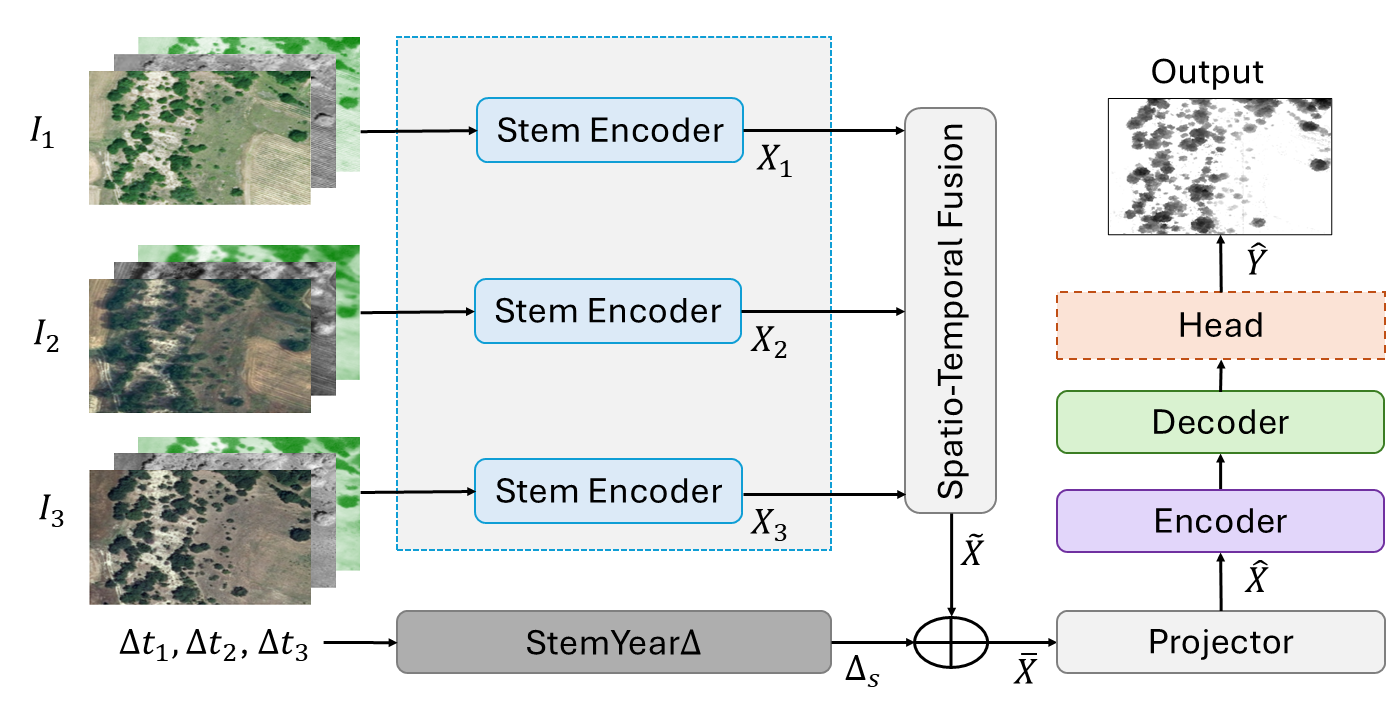}
    \caption{Flowchart of the proposed framework, taking three timestamp input images ($I_{i}$) where $i \in [1, 2, 3]$. Each $I_{i}$ input consists of RGB, NIR, and NDVI bands. These input images are input to the stem encoders to encode spatial stem features ($X_i$) for each timestamp. These spatial stem features are fused using the Spatio-Temporal Fusion block to obtain fused spatial-temporal embeddings ($\widetilde{X}$). We also input the relative time difference ($\Delta t_{i}$) in years, between the CHM timestamp (target) and each image acquisition date embeddings, to obtain StemYear$\Delta$ embeddings $\Delta_s$. These embeddings, along with  spatial-temporal embeddings ($\widetilde{X}$), are fused and input to a projector layer, which yields $\hat{X}$. These fused features are input to an encoder-decoder architecture and a head layer to obtain the final output $\hat{Y}$ as a canopy height map.}
    \vspace{-1em}
    \label{fig:main_network_diagram}
\end{figure}
\vspace{-0.8em}
\section{Canopy height prediction model}
\label{sec:model}

To implement this data-driven approach, we use a deep learning model specifically designed to predict canopy height from multi-temporal and multi-spectral imagery with geospatial metadata.\\ 
As shown in Fig. \ref{fig:main_network_diagram}, the model inputs three timestamps information ($I_{i}$, where $i\in [1,2,3]$ of size $256\times 256$ images), each containing five channels (RGB, NIR, and NDVI). In addition, yearly temporal information is provided through a vector of $\Delta t_i$ values, each representing the time difference in years between the CHM timestamp (target) and each image acquisition date. Suppose the CHM, the target $Y$, is acquired at time $t_y$, then $\Delta t_1 = t_y - t_1$, $\Delta t_2 = t_y - t_2$, and $\Delta t_3 = t_y - t_3$. \\ 
This architecture encodes the spatial stem features ($X_{i}$) for each input $I_{i}$ using the Stem Encoder block having two Conv–BN–ReLU layers (kernel size 3) at each timestamp $i$.  These spatial stem features ($X_i$) are fused using a spatio-temporal fusion block, where these spatial stem feature embeddings are concatenated and input to a projection layer to generate aggregated spatio-temporal embeddings $\widetilde{X}$.
This spatio-temporal processing handles multi-temporal RGB, NIR, and NDVI imagery through a spatio-temporal fusion block, extracting features while preserving both spatial and temporal information.\\
This model encodes relative time rather than absolute timestamps, providing temporal flexibility. With no hard-coded geographic parameters, the same methodology can be applied from small study areas to entire departments within France's national territory.
To do so, we also compute relative time difference embeddings $\Delta_s$ using $\Delta t_1, \Delta t_2$, and $\Delta t_3$ (the time difference in years between the CHM timestamp (target) and each image acquisition date)  to encode the the relative time difference among three timestamp images, which are crucial for tree prediction. These $\Delta_s$ embeddings are upsampled along spatial dimensions to match the spatio-temporal embeddings before fusion in spatial dimensions and fused with spatio-temporal embeddings $\widetilde{X}$ to obtain $\Bar{X}$, which encodes both spatio-temporal and relative time difference embeddings. These embeddings are realized using a projector layer, which yields $\hat{X}$, and forwarded to an encoder-decoder architecture to learn the spatio-temporal features and passed to a header to obtain the final canopy height map $\hat{Y}$.\\ 
\begin{figure}[t]
\vspace{-0.5cm}
\centering
\includegraphics[width=\textwidth]{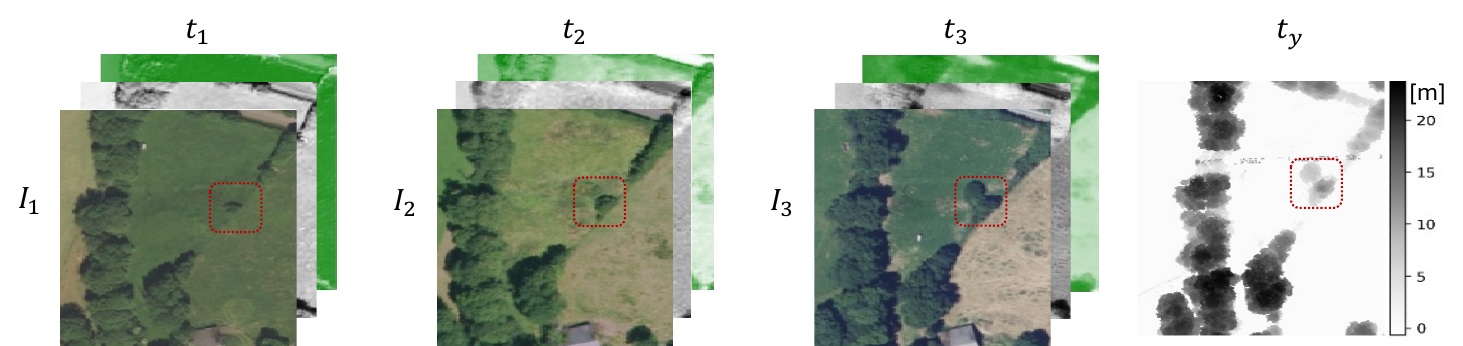}
\caption{A sample with visible tree growth, outlined in red, from PrediTree. We collect multi-spectral (RGB+NIR+NDVI) data at three timestamps $t_1, t_2, t_3$ and their corresponding spatially aligned CHM tile at $t_y$.}
\vspace{-0.5em}
\label{fig:timestamps}
\end{figure}
To address the specific challenges of vegetation height prediction, we implemented a specialized loss function. A masked mean squared error (MSE) loss focuses training on valid pixels while amplifying errors for vegetation, and does not collapse to a model only predicting $0$. Suppose  the $\hat{y}$ is the predicted output of the model, $y$ represents the ground truth, $m$ denotes a binary mask indicating valid spatial locations, and $w$ represents an amplification factor used to up-weight the loss where the target is greater than a threshold $\theta$. 
The equation for the weighted masked MSE follows:
\begin{equation}
\mathcal{L}_{\text{wmse}} (\hat{Y}) = \frac{ \sum_{i=1}^{N_\mathsf{x}} \sum_{j=1}^{N_\mathsf{y}} \left[ m_{i,j} w_{i,j} (\hat{y}_{i,j} - y_{i,j})^2 \right] }{ \sum_{i=1}^{N_\mathsf{x}} \sum_{j=1}^{N_\mathsf{y}} m_{i,j} },
\end{equation}
where $w_{i,j}$ is defined as $w=1+k\mathds{1}(y>\theta)$.
We set the constants $k$ and $\theta$ to 10 and 0.5m, respectively.
During inference, we compute the (unweighted) masked MSE and masked mean absolute error (MAE) errors.
The masked mean absolute error (MAE) during testing process is computed using the equation as:
\begin{equation}
\mathcal{L}_{\text{mmae}} (\hat{Y}) = \frac{ \sum_{i=1}^{N_\mathsf{x}} \sum_{j=1}^{N_\mathsf{y}} \left[ m^\prime_{i,j} |\hat{y}_{i,j} - y_{i,j}| \right] }{ \sum_{i=1}^{N_\mathsf{x}} \sum_{j=1}^{N_\mathsf{y}} m^\prime_{i,j} },
\end{equation}
where $m'=m \mathds{1}(y>\theta)$. In this study, the quantitative analysis of the regression performance is limited to the parts of the images with trees. We will present a few qualitative results of the detection problem, that is identifying which parts of the image contain trees or not.




\section{Experiments and results}
We perform a series of experiments using the proposed framework as described in \cref{sec:model}. 
These experiments have a two-fold objective: 1) to utilize the dataset in a canopy height prediction scenario and 2) to assess the contribution of different modalities, in this case spectral bands and temporal information, to prediction accuracy.

\vspace{-0.8cm}
\begin{figure}[ht!]
\centering
\includegraphics[width=\textwidth]{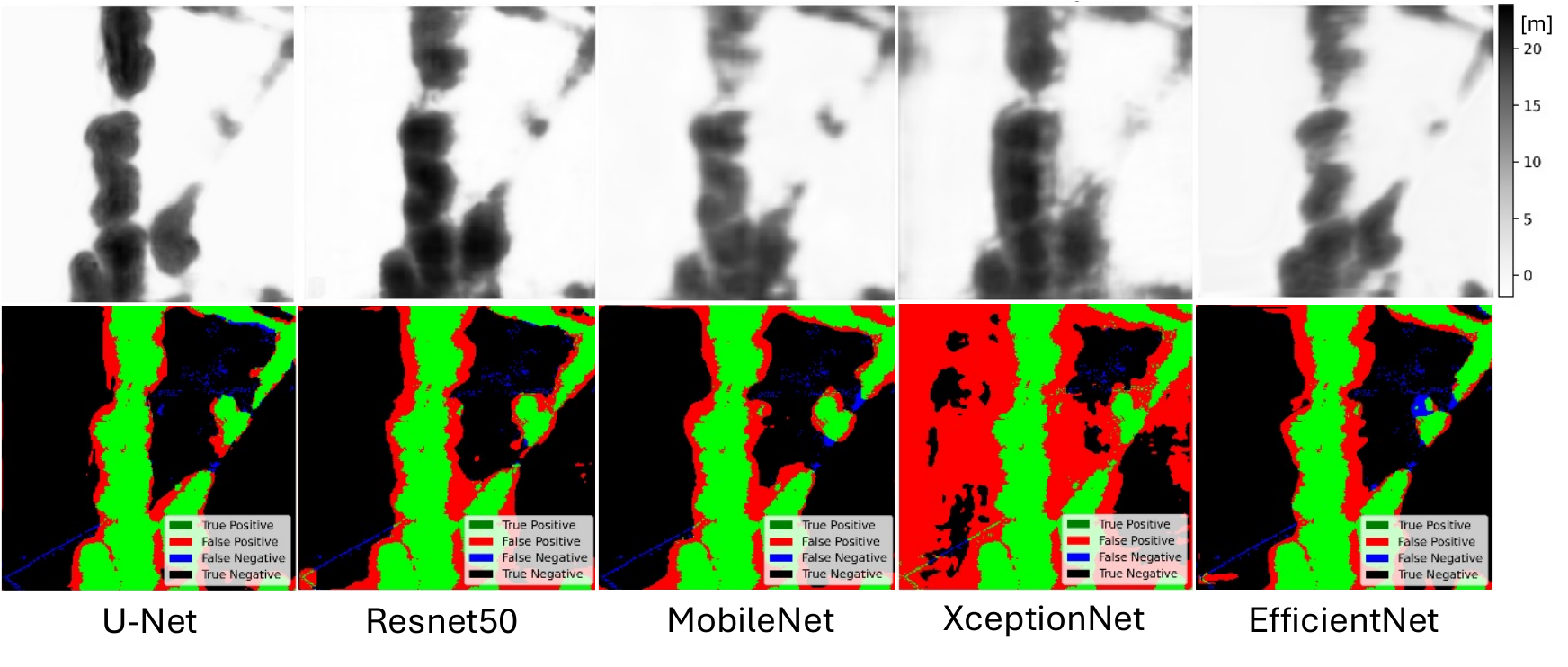}
\caption{Illustration of the CHM predictions for the sample in \cref{fig:timestamps}. In the first row, using different encoder-decoder architectures including: U-Net, Resnet50, MobileNet, XceptionNet and EfficientNet. The second row represents, w.r.t the CHM prediction and ground truth, true positives in \textcolor{green}{green}, false positives in \textcolor{red}{red}, false negatives in \textcolor{blue}{blue} and true negatives in black.}
\label{fig:architectures}
\end{figure}
\vspace{-1cm}
\begin{wraptable}[8]{r}{0.35\textwidth}
\vspace{-2.5em}
  \setlength{\belowcaptionskip}{2pt}    
  \setlength{\intextsep}{2pt}           
  \centering
  \scalebox{0.6}{
    \begin{tabular}{@{}c l l c c@{}}
      \toprule
      \textbf{Ex.} & \textbf{Encoder} & \textbf{Decoder} &
      \textbf{MSE} & \textbf{MAE}\\
      \midrule
      1 & U-Net        & U-Net      & \textbf{6.47} & \textbf{11.78}\\
      2 & ResNet-50    & DeepLabv3  & 7.37 & 11.95\\
      3 & MobileNet    & DeepLabv3  & 8.26 & 13.34\\
      4 & XceptionNet  & DeepLabv3  & 10.28 & 12.96\\
      5 & EfficientNet & DeepLabv3  & 8.69 & 13.24\\
      \bottomrule
    \end{tabular}
  }
  \caption{Quantitative comparison of encoder–decoder architectures. Lower is better.}
  \label{tab:main_quantitative}
\end{wraptable}           

\subsection{Experimental setup}
\label{ssec:exp}
We implemented the experiments using PyTorch and trained all models on a single NVIDIA A100 GPU with 40GB RAM. 
The dataset consists of 785,392 samples ($\approx 800$GB) with four images each, which, if used in its entirety, is quite challenging for training. Therefore we used a subset of the entire dataset consisting of 30K samples as a training set and 393 samples as the test set. During training, it took about 1 hour to complete 1 epoch using 30K samples.
The dataset consists of 3 images (RGB, NIR, and NDVI at 3 different timestamps, and 1 CHM) at 0.5m resolution, each covering a $256\times 256$ pixel area. \\
For model training, we used the AdamW optimizer with an initial learning rate of $1e-4$ and a weight decay of 0.01. We implemented a learning rate reduction strategy that decreased the learning rate by a factor of 0.5 when validation loss plateaued for 5 epochs. All models were trained for 100 epochs with a batch size of 32.
To evaluate model performance, we used complementary metrics, the Mean Squared Error (MSE), for overall height prediction accuracy, and the Mean Absolute Error (MAE) for height prediction accuracy. 
\subsection{Quantitative and Qualitative Analysis}
In \cref{tab:main_quantitative}, we utilized different state-of-the-art encoder-decoder architectures in the canopy height prediction framework and compared their performance for tree canopy height estimation. 
To do so, we remove their stem layers for each network and use their encoder-decoder architecture.
We set U-Net \cite{ronneberger2015u} architecture as the baseline framework. For the rest of the networks, we change the encoder to ResNet50 \cite{he2016deep}, MobileNet \cite{sinha2019thin}, Xception \cite{chollet2017xception}, EfficientNet \cite{tan2019efficientnet}, and keep the same decoder as Deeplabv3 \cite{chen2017rethinking}. We notice that the experiments 3 and 5 using the EfficientNet and MobileNet encoders have more than 8\% MSE and 13\% MAE scores. Whereas employing U-Net  \cite{ronneberger2015u} into the framework results in better MSE and MAE scores. \\
In \cref{fig:timestamps}, the first row displays a test sample, showing (from left to right): the three timestamp inputs $I_i$ having 5 bands (RGB, NIR, and NDVI) and the ground truth height map. In \cref{fig:architectures}, the first row represents the predicted height map, and the last row indicates the color-coded tree detection, with true positives, false positives, false negatives and true negatives. \\
We compared different encoder-decoder-based architectures employed in the tree prediction height framework. We observe that U-Net performs better compared to other models, where height predictions remain more accurate despite the dense forest and complex surroundings.

\begin{wraptable}{r}{0.45\textwidth}
\vspace{-1em} 
    \setlength{\belowcaptionskip}{2pt}    
    \setlength{\intextsep}{2pt}  
\centering
\scalebox{0.7}{%
\begin{tabular}{@{}c c c c c c c@{}}
\toprule
\textbf{Ex. No.} & \textbf{RGB} & \textbf{NIR} & \textbf{NDVI} & \textbf{Timestamps} & \textbf{MSE} & \textbf{MAE} \\
\midrule
1 & \textcolor{brightgreen}{\checkmark} & \textcolor{red}{\ding{55}} & \textcolor{red}{\ding{55}} & $t_1, t_2, t_3$   & 9.23 & 13.47 \\
2 & \textcolor{red}{\ding{55}} & \textcolor{brightgreen}{\checkmark}  & \textcolor{red}{\ding{55}} & $t_1, t_2, t_3$   & 10.68 & 13.61 \\
3 & \textcolor{red}{\ding{55}} & \textcolor{red}{\ding{55}} & \textcolor{brightgreen}{\checkmark}  & $t_1, t_2, t_3$  & 12.35 & 13.67 \\
4 & \textcolor{brightgreen}{\checkmark} & \textcolor{brightgreen}{\checkmark}  & \textcolor{red}{\ding{55}} & $t_1, t_2, t_3$   & 8.68 & 13.34 \\
5 & \textcolor{brightgreen}{\checkmark} & \textcolor{red}{\ding{55}} & \textcolor{brightgreen}{\checkmark}  & $t_1, t_2, t_3$   & 9.00 & 13.55 \\
6 & \textcolor{red}{\ding{55}} & \textcolor{brightgreen}{\checkmark}  & \textcolor{brightgreen}{\checkmark}  & $t_1, t_2, t_3$   & 9.84 & 13.51 \\
7 & \textcolor{brightgreen}{\checkmark} & \textcolor{brightgreen}{\checkmark}  & \textcolor{brightgreen}{\checkmark}  & $t_3$       & 9.29 & 13.25 \\
8 & \textcolor{brightgreen}{\checkmark} & \textcolor{brightgreen}{\checkmark}  & \textcolor{brightgreen}{\checkmark}  & $t_2, t_3$     & 7.79 & 12.66 \\
9 & \textcolor{brightgreen}{\checkmark} & \textcolor{brightgreen}{\checkmark}  & \textcolor{brightgreen}{\checkmark}  & $t_1, t_2, t_3$   & \textbf{6.47} & \textbf{11.78} \\
\bottomrule
\end{tabular}%
}
\caption{Ablation study evaluating the impact of input modalities and timestamps. Best results are in bold.}
\label{tab:ablation_results}
\end{wraptable}
\subsection{Ablations studies}
\label{ssec:ablation}
To understand the relative importance of different input modalities 
, we conducted an extensive ablation study examining both spectral and temporal dimensions. 
In Table \ref{tab:ablation_results}, we utilize various spectral bands to validate their effectiveness. First, we employ separately RGB, NIR, and NDVI in the framework, and notice that only the RGB bands give comparatively better 9.23\% MSE and 13.47\% MAE scores. After that, we apply different combinations of using only two modalities and found that utilizing RGB and NIR results in a better score, as shown in Exp. No. 4. It is evident in Exp. No. 9 that utilizes all spectral bands, including RGB, NIR, and NDVI, results in the best performing scores, such as 6.47\% MSE and 11.78\% MAE. Although the NDVI provides redundant information with the NIR and R bands, we observe that giving both outperforms giving only the NIR (Exp. No. 2 vs. Exp. No. 6). We believe that this is because providing the NDVI from the start to the network helps the model learn faster vegetation-related patterns, which yields a better performance after a fixed number of epochs. We deduce that, the more the modalities, the better the predictions. 
The temporal ablation experiments reveal that using all three timestamps consistently yielded the best results, as shown in Exp. No. 9 in Table \ref{tab:ablation_results}. 

\section{Conclusion and discussion}
We presented PrediTree, the first openly available, sub-meter, multi-temporal and multi-spectral CHM dataset tailored for training models for predicting tree height. It is composed of 3,141,568 $256\times256$ pixel tiles at 0.5m resolution with spatially aligned LiDAR-derived canopy height maps. We propose an encoder-decoder framework to handle the PrediTree dataset and predict the tree growth for the future. The detailed experiments validate that the framework performs better in predicting the tree canopy height maps. We provide openly available codes that allow users to tailor this dataset to their needs, generating custom datasets at any resolution from 0.15m, 0.2m or 0.5m upwards. This scalable framework can process data from almost 80 departments across France using the same standardized format as the benchmark dataset. While PrediTree currently focuses on French forests, the framework is designed for global expansion. Future releases will include 1) more data modalities such as climate data and soil data, as we have illustrated that adding more modalities contributed to increasing performance, to bridge the gap with 3-PG, 2) additional countries to add variability to the current dataset and finally, 3) extended temporal sequences, \textit{i.e.,} multiple $\Delta t$ values.
%
%
%
\bibliographystyle{splncs04}
\bibliography{mybibliography}
%




\end{document}